\documentclass[journal]{IEEEtran}
\usepackage{cite}
\usepackage{multirow} 

\usepackage{color,xcolor}
\definecolor{mod_01}{rgb}{0,0,1}

\ifCLASSINFOpdf
  \usepackage[pdftex]{graphicx}
  \graphicspath{{../pdf/}{../jpeg/}}
  \DeclareGraphicsExtensions{.pdf,.jpeg,.png}
\else
\fi
\usepackage{amsmath}
\begin{document}
%
\title{Uncovering Dominant Features in Short-term Power Load Forecasting Based on Multi-source Feature}
%
%
%

\author{Pan~Zeng$^*$,
        and~Md~Fazla~Elahe,
        Junlin~Xu,
        Min~Jin
\thanks{\emph{Corresponding author: Pan Zeng.}}
}

\maketitle

\begin{abstract}
Due to the limitation of data availability, traditional power load forecasting methods focus more on studying the load variation pattern and the influence of only a few factors such as temperature and holidays, which fail to reveal the inner mechanism of load variation. 
This paper breaks the limitation and collects 80 potential features from astronomy, geography, and society to study the complex nexus between power load variation and influence factors, based on which a short-term power load forecasting method is proposed.
Case studies show that, compared with the state-of-the-art methods, the proposed method improves the forecasting accuracy by 33.0\% to 34.7\%.
The forecasting result reveals that geographical features have the most significant impact on improving the load forecasting accuracy, in which temperature is the dominant feature. Astronomical features have more significant influence than social features and features related to the sun play an important role, which are obviously ignored in previous research. Saturday and Monday are the most important social features. Temperature, solar zenith angle, civil twilight duration, and lagged clear sky global horizontal irradiance have a V-shape relationship with power load, indicating that there exist balance points for them. Global horizontal irradiance is negatively related to power load.

\end{abstract}


%
\IEEEpeerreviewmaketitle

\section{Introduction}
%
%
%
%
\IEEEPARstart{W}{ith} the continuous development of renewable energy and the diversity of power demand in the energy market, the nonlinearity and randomness of power load are becoming more obvious. As a result, short-term load forecasting has become a very challenging study. There are various factors that affect the load variation and the relationship between them is complicated. Selecting proper features and studying the complex nexus between load variation and influence factors become the keys to improve the forecasting accuracy.

In recent years, short-term load forecasting is experiencing an important change from solely studying the variation pattern of power load to exploring the key factors that cause the load fluctuation\cite{Hverstad2015short,xie2016relative}. Weather is considered as the most important factor that affects the power load. Reference \cite{wi2011holiday} analyzes the correlation between weather factors and electric load with mutual information and believes that discomfort index and temperature are the dominant weather factors that affect the load variation of holidays. Reference \cite{wang2016electric} explores the influence of lagged hourly temperature and moving averaging temperature on load forecasting, and develops a forecasting method with better forecasting accuracy. 
On the one hand, weather factors have been proved to be significantly related to power load variation\cite{zeng2017peak,son2017short}.
On the other hand, comprehensive historical and forecast weather data of almost everywhere in the world are available to the public, making it possible for researchers to study load forecasting based on weather data. Related research shows that weather is not the only factor that causes load variation. 
Aiming at improving the forecasting accuracy of holidays, \cite{pan2019learning} proposes a method based on transfer learning and improves the forecasting performance. This research believes that the load variation pattern of holidays is significantly different than that of non-holidays. Reference \cite{guo2018deep} reveals that air-quality-related factors affect human engagement in outdoor activities and thus alter load variation patterns. Reference \cite{moral2017integrating} proposes a forecasting procedure based on GDP, GVA, consumption, etc., which reflects the influence of economic factors on load variation.
However, these factors are only part of the causes of load variation, and with limited factors, the inner mechanism of load variation could not be fully revealed.
Individual and social power consumption regularity indicates that many factors such as solar irradiance, NO$_{2}$ content, and tide, may also have potential correlations with load variation, however, there are limited related studies. The main reason is that these data were difficult to obtain in the past. In recent years, with the development of the Internet and the establishment of various public data platforms, more and more data are available for researchers.
For example, the National Renewable Energy Laboratory\footnote{https://www.nrel.gov/} provides solar irradiance data, geothermal data, etc., and the United States Environmental Protection Agency\footnote{https://www.epa.gov/} provides varieties of atmospheric data including nitrogen oxide content and air quality index.
The opening of these platforms has greatly expanded the types and volumes of datasets related to load forecasting and provides strong support for systematically exploring the inner mechanism of load variation. 

In terms of model construction, various load forecasting models have been proposed, such as support vector regression (SVR)\cite{Jiang2018Ashort}, random forest, etc. In practical application, the performance of a single model varies when applied to different datasets, and different models may lead to different accuracy when applied to the same dataset. In order to make up the deficiencies of single model methods, multi-model methods are proposed\cite{Wang2011ELITE,Wang2018an}. The fundamental idea of multi-model is to take the advantages of different single models thus archiving higher forecasting accuracy. Reference \cite{nowotarski2016improving} constructs a set of models by different subsets of feature variables, combines the results of them, and improves the forecasting accuracy. Reference \cite{zhou2017holographic} proposes a holographic ensemble forecasting method, which constructs multiple training sets by performing diversity sampling and generate multi-models with multiple algorithms, and obtains better forecasting performance.

Although various forecasting methods have been proposed, the main objective of much research is improving the forecasting performance, and they usually do not focus on studying the inner mechanism of load variation and the complex factor-load nexus. Methods like multiple linear regression, while interpretable, fail to obtain high forecasting performance. Methods based on intelligent algorithms focus on improving the forecasting performance whereas they always fail to study the interpretability of forecasting models. Aiming at solving this problem, in this paper, we select feature variables based on multi-source data and construct multi-source feature (MSF), which collect up to 80 potential features from astronomy, geography, and society in order to fully reveal the inner mechanism of load variation and the complex factor-load nexus. The aim of this research is to (a) propose an accurate forecasting model base on MSF, (b) uncover the dominant features that have the most significant influence on forecasting accuracy, and (c) further study the complex nexus between dominant factors and load variation patterns.

The rest of this paper is organized as follows. Section II introduces the feature selection method based on MSF and describes the datasets used in this paper. Section III presents the case studies and discusses the performance of different feature selection scenarios. Section VI discusses the importance of different dominant features and the correlation between dominant factors and load variation. Section V concludes this research.


\section{Method and Dataset}

\subsection{Feature Seletion Method Based on MSF}
There are various factors that cause the load variation, which can usually be divided into two categories: natural factors and social factors\cite{zeng2017peak}. Reference \cite{jingrui2018load} replaces the month attribute by traditional Chinese solar terms as the date attribute for load forecasting and achieves better accuracy, which implies that the positional relationship between the sun and the earth has a non-negligible impact on the power load, indicating that the load variation may be related to astronomical factors.
On the one hand, in order to deeply explore the inner mechanism of load variation, reveal the complex factor-load nexus, and finally improve the forecast accuracy, it is necessary to consider as many related factors as possible. On the other hand, due to the accessibility of public data platforms and research institutes in different fields, a large amount of astronomical, geographical, and social data are available for researchers, which makes it possible to carry out load forecasting research based on these data. 
Therefore, this study constructs MSF by selecting feature variables from three aspects: astronomy, geography, and society. Among them, astronomical factors (A-factors) include global horizontal irradiance (GHI),  clear sky GHI (CKGHI), moon phase, tide, etc. Geographical factors (G-factors) include temperature, air pressure, air quality, etc. Social factors (S-factors) include holidays, weekdays, etc.
The MSF and historical load data constitute the candidate feature dataset, as describe in (\ref{eq_00})

\begin{equation}
X=[G_1,\ldots,G_i,A_1,\ldots,A_j,S_1,\ldots,S_k,L_1,\ldots,L_l]
\label{eq_00}
\end{equation}

where,\emph{G,A,S,L} represent geographical features (G-features), astronomical features (A-features), social features (S-features), and historical load features, respectively. \emph{i,j,k,l} represent the number of features in each groub.

\subsection{Feature Selection}
The feature selection method implemented in this research includes two steps. First, the variance of each feature variable is calculated, and if it is less than a threshold, the corresponding feature would be removed\footnote{https://scikit-learn.org/stable/modules/generated/sklearn.feature\_selection.\\VarianceThreshold}. In this research, the threshold is 0.1056. In the second step, assuming there are \emph{n} samples, the correlation between the \emph{i}th feature and the label is computed, as in
\begin{equation}
{r_i} = \frac{{\sum\nolimits_{j = 1}^n {({X_{ij}} - \bar X)({y_j} - \bar y)} }}{{\sqrt {\sum\nolimits_{j = 1}^n {{{({X_{ij}} - \bar X)}^2}}  \cdot \sum\nolimits_{j = 1}^n {{{({y_j} - \bar y)}^2}} } }}
\label{eq_04}
\end{equation}

Then, the score for the \emph{i}th feature \emph{f$_{i}$} is calculated according to (\ref{eq_05}).
\begin{equation}
{f_i} = \frac{{r_i^2}}{{1 - r_i^2}}(n - 2)
\label{eq_05}
\end{equation}

 Finally, top \emph{k}th features are selected for modeling\footnote{https://scikit-learn.org/stable/modules/generated/sklearn.feature\_selection.\\SelectKBest.html}. This feature selection method is referred to as the LV-KB method in this paper.

\subsection{Benchmark Models and Evaluation Metrics}
In this paper, we construct forecasting models with three commonly used algorithms, namely support vector regression (SVR), gradient boosting regression tree (GBRT), and multilayer perceptron (MLP). SVR provides satisfactory accuracy but it is sensitive to outliers. GBRT is more robust to outliers. MLP has a strong nonlinear learning ability whereas it is easy to overfit when the training set is small. 
The kernel function of SVR is \emph{Linear} and the loss function of GBRT is least squares regression. For MLP, the solver for weight optimization is \emph{lbfgs}, the number of the hidden layer is 2 and the number of nodes in two hidden layers are 5 and 2. Other parameters are determined by the grid search method.

Mean absolute error (MAE), mean absolute percentage error (MAPE), and root mean squared error (RMSE) are used to evaluate the forecasting performance, as shown in (\ref{eq_02}), (\ref{eq_06}), (\ref{eq_07}).

\begin{equation}
{\emph{MAE} = \frac{1}{N}\sum\limits_{t = 1}^N {\left| {{{\hat y}_t} - {y_t}} \right|} }
\label{eq_02}
\end{equation}

\begin{equation}
{\emph{MAPE} = \frac{1}{N}\sum\limits_{t = 1}^N {\left| {\frac{{{{\hat y}_t} - {y_t}}}{{{y_t}}}} \right|}  \times 100\% }
\label{eq_06}
\end{equation}

\begin{equation}
{\emph{RMSE} = \sqrt {\frac{1}{N}\sum\limits_{t = 1}^N {{{({{\hat y}_t} - {y_t})}^2}} }}
\label{eq_07}
\end{equation}

\subsection{Dataset Description}
In order to study the factor-load nexus of different regions, we use three public datasets to conduct case studies, namely Maine dataset from ISO New England electricity market\footnote{https://www.iso-ne.com/isoexpress/web/reports/load-and-demand/-/tree/zone-info}, New South Wales dataset from Australian Energy Market Operator\footnote{https://aemo.com.au/en/energy-systems/electricity/national-electricity-market-nem/data-nem/aggregated-data}, and Texas dataset from The Electric Reliability Council of Texas\footnote{http://www.ercot.com/gridinfo/load/load\_hist}. 

\subsubsection{Maine Dataset}
\begin{figure}[!t]
\centering
\includegraphics[width = 3.2in]{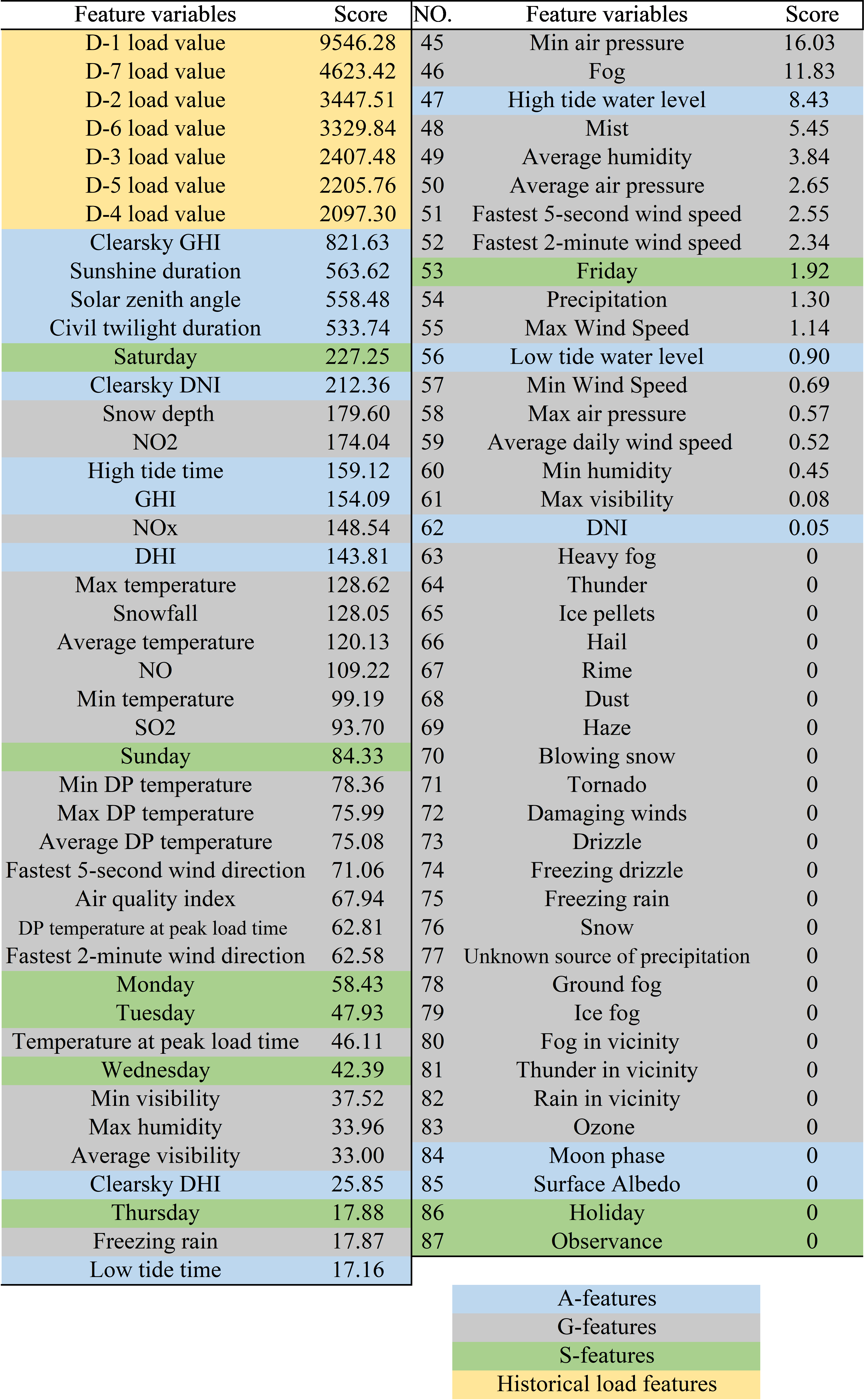}
\caption{MSF of Maine dataset}
\label{fig_01}
\end{figure}

The data of Maine are daily peak load from 2003 to 2015 and we collect 80 candidate feature variables to construct MSF, including 56 G-features, 15 A-features, and 9 S-features, as shown in Fig.\ref{fig_01}. Features with a score of 0 represent that they are removed in the first step of LV-KB.

\subsubsection{NSW Dataset}
\begin{figure}[!t]
\centering
\includegraphics[width = 3.2in]{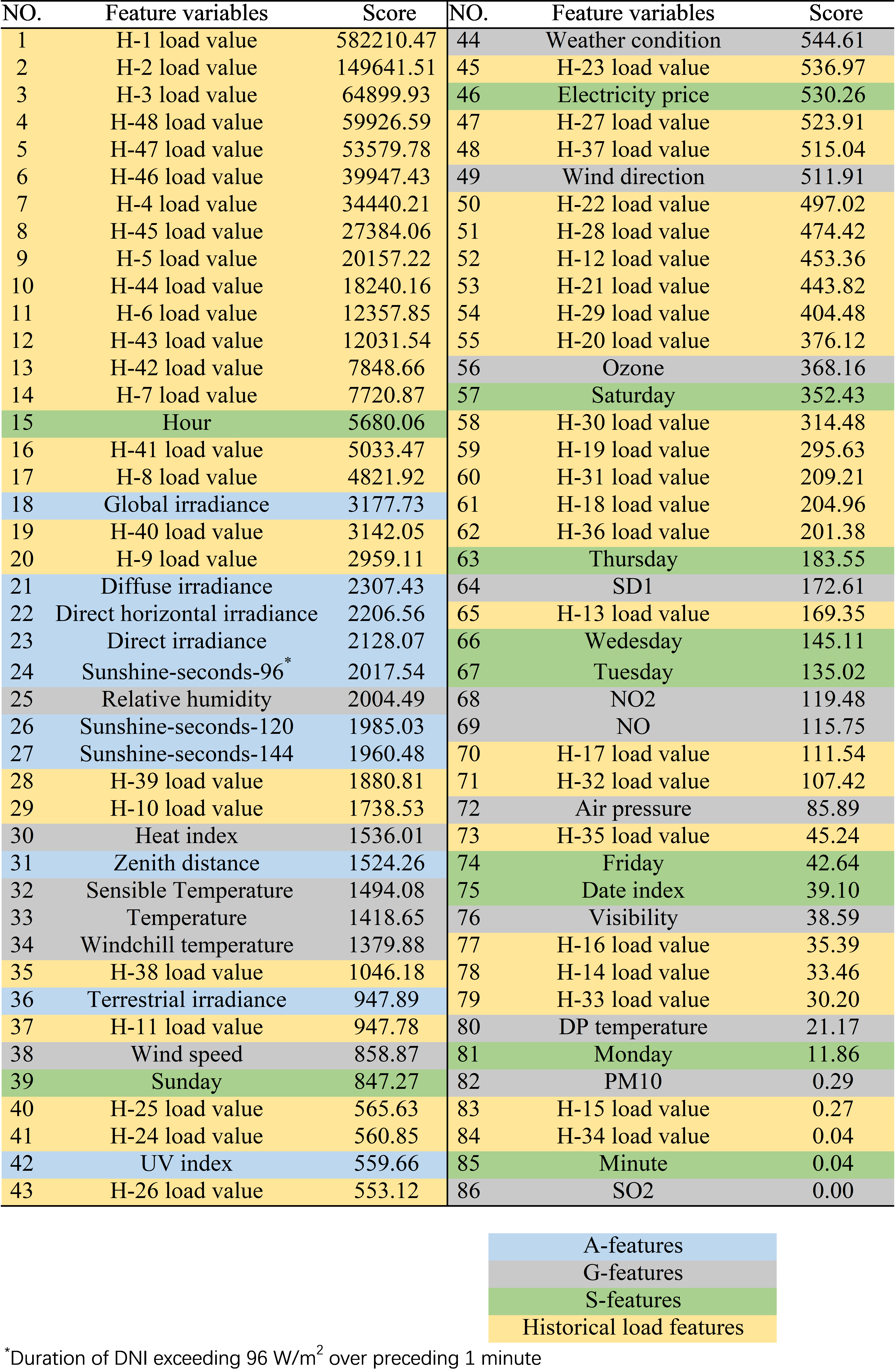}
\caption{MSF of NSW dataset}
\label{fig_02}
\end{figure}

The data of NSW are half-hourly load from January 1, 2009 to January 6, 2010, and the candidate feature variables used in this dataset is presented in Fig.\ref{fig_02}.

\subsubsection{NSW Dataset}
\begin{figure}[!t]
\centering
\includegraphics[width = 3.0in]{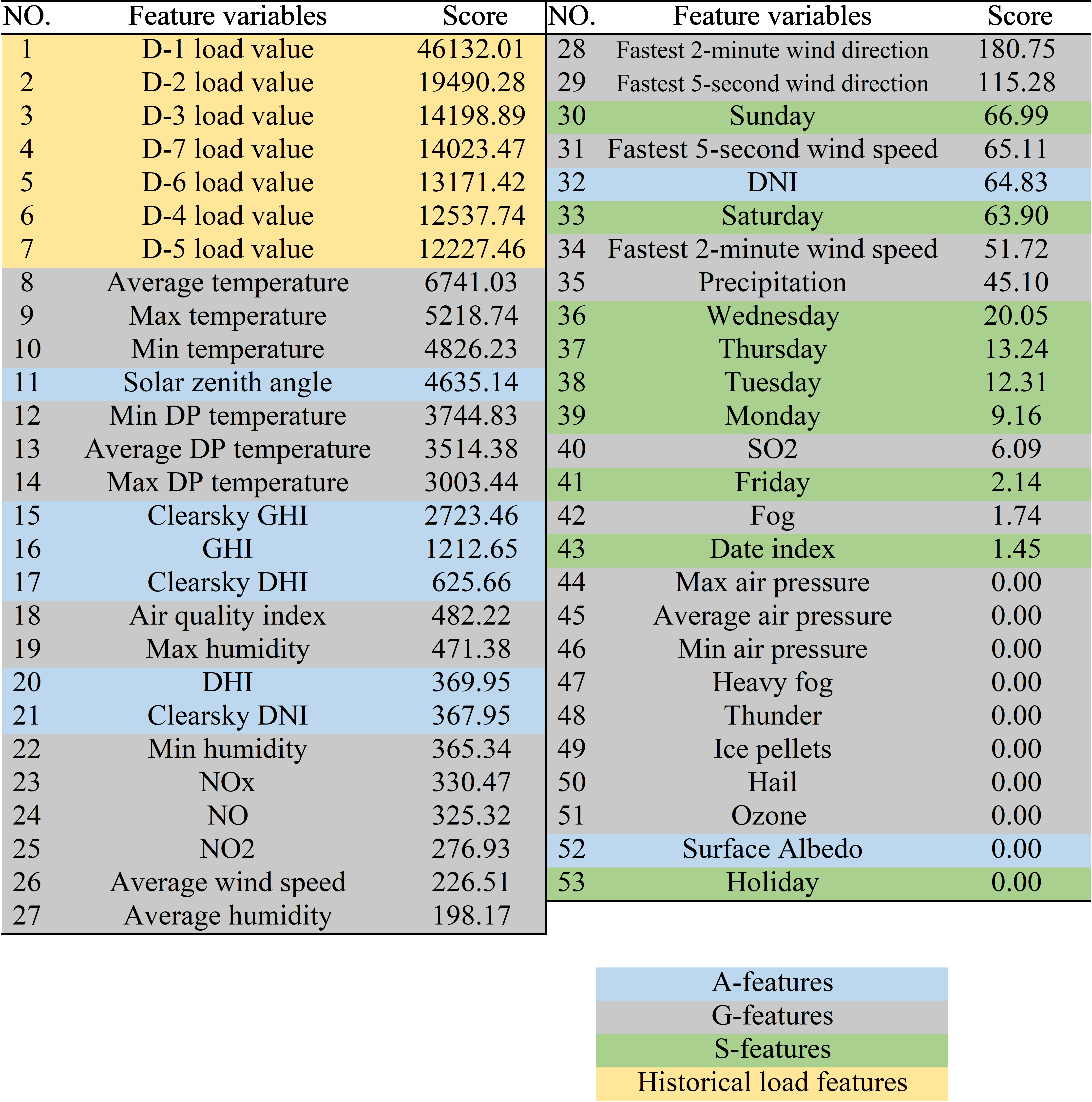}
\caption{MSF of Texas dataset}
\label{fig_03}
\end{figure}

The data of Texas are daily peak load from 1998 to 2017, and the candidate feature variables used in this dataset is presented in Fig.\ref{fig_03}.

\section{Case Studies}

\subsection{Case I: Maine Dataset}
\begin{figure}[!t]
\centering
\includegraphics[width = 2.3in]{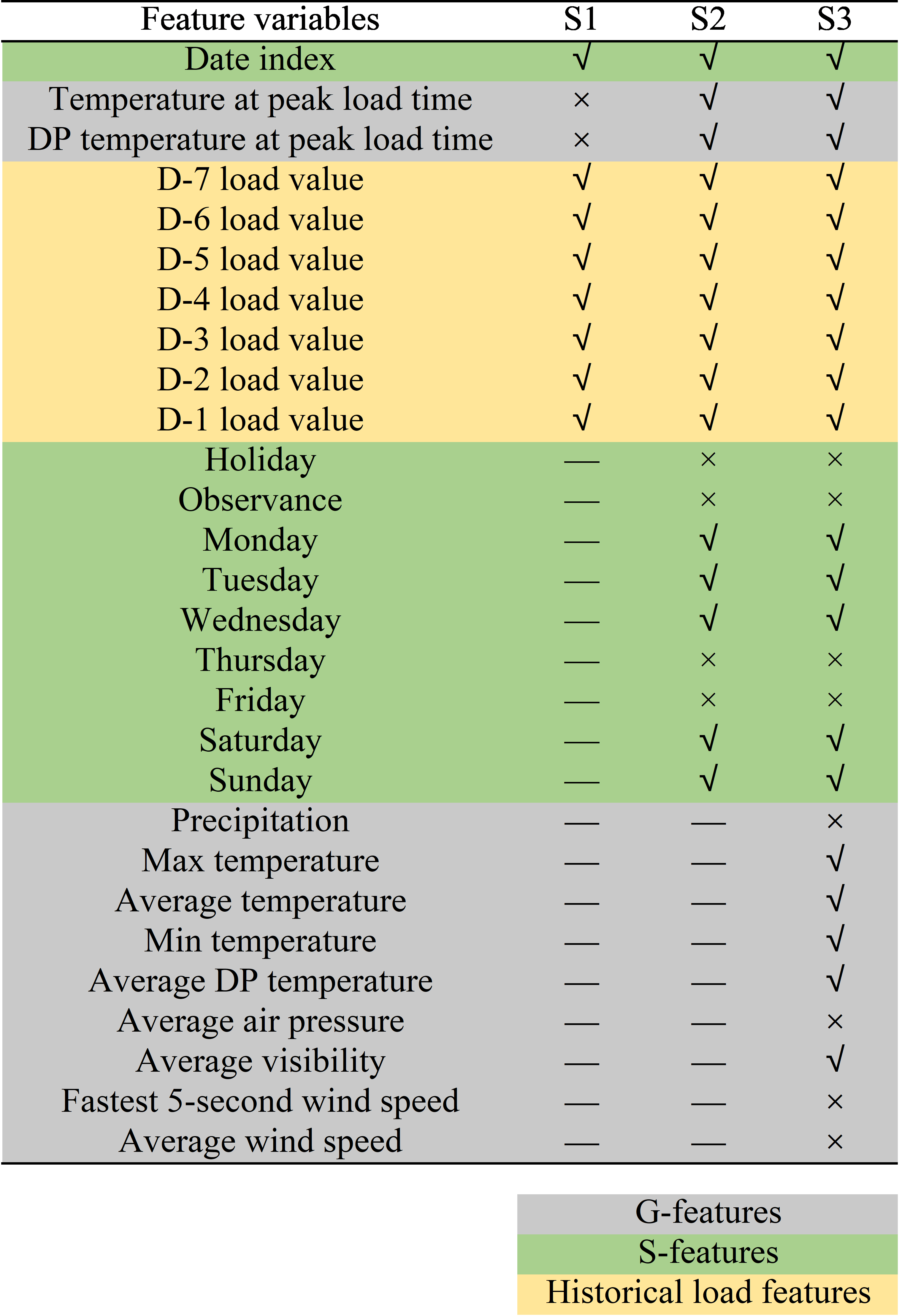}
\caption{Candidate feature variables of S1, S2, and S3}
\label{fig_04}
\end{figure}

In this case, we compare four different feature scenarios with Maine dataset. The first three scenarios are shown in Fig.\ref{fig_04}. S1 selects temperature and dew point temperature at the peak load time and the load of the last 7 days as the candidate features. S2 adds 9 S-features based on S1; S3 adds 9 G-features selected by \cite{alipour2019assessing}. In Fig.\ref{fig_04}, the sign of $\surd$ and $\times$ represent the candidate features in each dataset, and the sign of $\surd$ represents the input features. The fourth scenario is based on MSF, which collects 80 candidate features and 7 historical load features, as shown in Fig.\ref{fig_01}. We apply LV-KB method to these four datasets and select 8, 15, 20, and 55 input features, respectively.
Data of 2015 are used for testing, and the rest are used as the training set. 

\begin{figure}[!t]
\centering
\includegraphics[width = 2.5in]{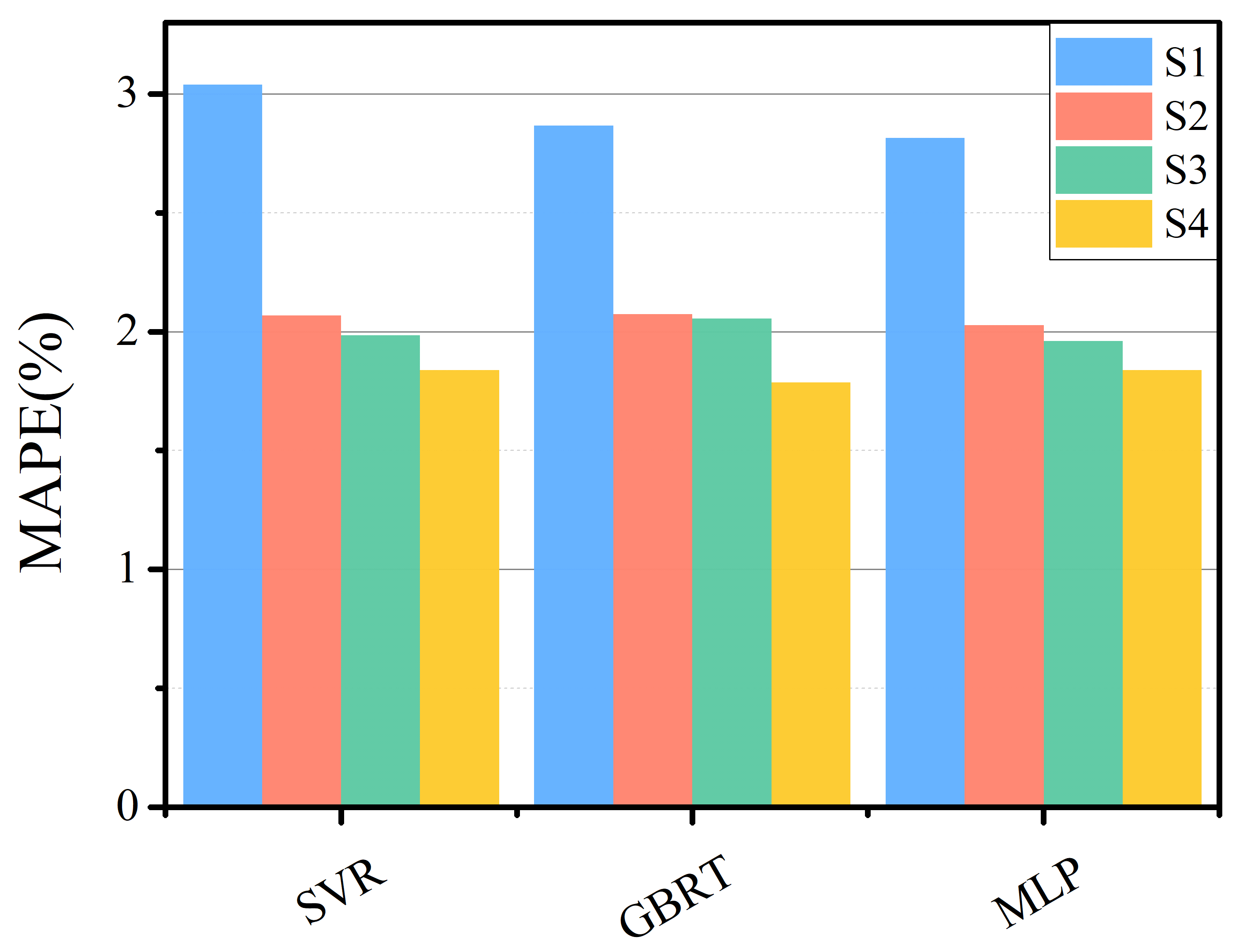}
\caption{Result of case I}
\label{fig_05}
\end{figure}

As shown in Fig.\ref{fig_05}, the accuracy of S1 is the lowest and that of S4 is the highest. The performance of S2 and S3 are very close to each other, whereas that of S3 is slightly better.
As presented in Fig.\ref{fig_04}, S1 only contains 7 historical load features and date index, whereas related studies have proved that temperature is one of the most important factors that affect the load variation. Therefore, the accuracy based on S1 is the worst. Compared with S1, S2 contains temperature, dew point temperature, and five weekday attributes, and the accuracy based on S2 is greatly improved. S3 contains 5 more features than S2, however, four of them are redundant with the temperature and dew point temperature included in S2. As a result, the improvement of accuracy brought by S3 is limited.

From the ranking of feature variables showed by Fig.\ref{fig_01} we can see that the first 7 features are the load value of the previous 7 days, followed by four A-features, namely CKGHI, sunshine duration, solar zenith angle (SZA), and civil twilight duration (CTD), showing a strong correlation between load variation and A-factors.

The feature scenario based on MSF (S4) contains more features from three aspects, especially the A-features that have been ignored by previous studies. According to the scores of features, there is a strong correlation between A-factors and load variation, therefore, S4 could obtain better forecasting accuracy.

\subsection{Case II: NSW Dataset}
Reference \cite{Wang2019Bi} proposes a load forecasting method based on attention mechanism, rolling update, and bi-directional long short-term memory neural network and obtains better forecasting performance. The case studies carried out by \cite{Wang2019Bi} are to forecast the half-hourly load of NSW from December 31, 2009, to January 6, 2010. Features used by \cite{Wang2019Bi} include historical load value, dry bulb temperature, dew point temperature, wet bulb temperature, humidity, and electricity price. 
To compare with this method, we collect 86 candidate feature variables based on multi-source data, as shown in Fig.\ref{fig_02}, and the LV-KB method is applied to select the top 55 features. Again, we use SVR, GBRT, and MLP to build forecasting models and the data from December 1 to December 30, 2009 are used as the training set. We run each experiment 30 times and use the average value of MAPE, RMSE, and time cost to compare with the results of \cite{Wang2019Bi}. Table \ref{tab_01} shows the comparison results.
\begin{table}[!t]
\centering
\caption{Comparison results of Case II.}
\label{tab_01}
\begin{tabular}{@{}cccc@{}}
\hline
\hline
Model                   & MAPE(\%)    & RMSE(mW)   & Time cost(s)    \\
\hline
Method proposed by \cite{Wang2019Bi} & 1.030 & 103.53 & 24.4  \\
SVR                                & 0.719 & 77.60  & 7.1 \\
GBRT                               & 1.032 & 107.46 & 3.8  \\
MLP                                & 0.798 & 85.25  & 13.8  \\
\hline
\hline
\end{tabular}
\end{table}

SVR, GBRT, and MLP are widely used machine learning algorithms, whereas the Bi-LSTM model is a deep learning method. The learning ability of deep learning method is better than that of SVR, GBRT, and MLP. However, as presented in Table \ref{tab_01}, not only the forecasting accuracy of SVR and MLP are higher than that of the method proposed by \cite{Wang2019Bi}, the time cost of them are also much lower. The performance of GBRT is close to the Bi-LSTM method whereas the time cost of GBRT is lower. Considering that the learning ability of GBRT is not as high as that of deep learning method, the comparison results still show the improvement brought by MSF.

It can be seen that even with single model methods, which usually have lower time cost, the application of MSF can still bring obvious improvement to the forecasting accuracy. If we combine MSF with multi-model or deep learning methods, the forecasting performance may be further improved.

\subsection{Case III: Texas Dataset}
Bayesian additive regression trees (BART) is a Bayesian sum-of-tree model\cite{chipman2010bart}. In short-term load forecasting, BART is considered to be an accuracy model which could effectively capture the nexus between load consumption and climate variability\cite{alipour2019assessing}. 
Reference \cite{alipour2019assessing} uses the daily peak load of Texas from 2002 to 2017 to conduct experiments. The feature variables used by \cite{alipour2019assessing} include average temperature, average dew point temperature, average sea level pressure, average visibility, average wind speed, maximum sustained wind speed, maximum temperature, minimum temperature, precipitation, per capita real gross state product, unemployment percentage, electricity price, etc. 
The out-of-sample model performance was estimated using a 20\% holdout cross-validation approach in \cite{alipour2019assessing} and the mean MSE and mean RMSE are calculated after 30 iterations. 
Following the same experimental procedure, we apply MSF to the BART method with 20\% holdout cross-validation approach and compare the results with \cite{alipour2019assessing} after 30 iterations. Table \ref{tab_02} shows the results.
\begin{table}[!t]
\centering
\caption{Comparison results of Case III}
\label{tab_02}
\begin{tabular}{@{}ccc@{}}
\hline
\hline
Model      & RMSE(gW)           & MAE(gW)              \\
\hline
BART proposed by \cite{moral2017integrating} & 2.866           & 2.213           \\
BART with MSF   & 1.920          & 1.444          \\
\hline
\hline
\end{tabular}
\end{table}

As shown in Table \ref{tab_02}, the forecasting performance of BART with MSF is better, which improves the RMSE and MAE by 33.0\% and 34.7\% respectively. From Fig.\ref{fig_03} we can see that MSF includes features from astronomical aspects and more feature variables in geographical aspects, therefore it contains more information of load variation patterns, and thus fundamentally improves the forecasting accuracy.

From case studies we can see that, first, compared with traditional methods based on natural and social features, MSF introduces the features from astronomical aspects for the first time, which contains diversity and large-scale candidate feature data. Comparative experiments show that the application of MSF significantly improves the forecasting accuracy. 
Second, compare the approach that builds an accurate but complex forecasting model with the approach that uses more features that are closely related to load variation, if the forecasting accuracy of them are close to each other, it's obvious that the latter approach could obtain lower time consumption and complexity. 
Last, since the three datasets are from different regions, they have completely different characteristics in weather, climate, residents’ living habits, geographical location, etc. The models used in case studies are also different from each other. The experiment results based on these datasets and models can fully illustrate that the approach based on MSF is dataset-independent and model-independent, showing that this method has a wide scope of applications and excellent generalization ability.

\section{Further Discussion}
In Section III, the effectiveness of MSF is fully illustrated by three case studies. However, these case studies only demonstrate that MSF could improve the forecasting performance, yet they have not explained how MSF affects the forecasting result. In this section, we try to discuss the following questions. Which kind of features have the most significant influence on load forecasting? Which are the dominant features? What is the relationship between dominant factors and power load variations?
In order to study these questions, the daily peak load of Maine from 2003 to 2014 is used as the training set and that of 2015 is used for testing.

\subsection{Uncovering the Dominant Features}
To study the importance of the features from three aspects, we use the top 10 G-features, top 10 A-features, and top 7 S-features for modeling. The corresponding features are combined with the historical load features as the input. The results are shown in Table \ref{tab_03}.
\begin{table}[!t]
\centering
\caption{Forecasting performance (MAPE) of features of different aspects.}
\label{tab_03}
\begin{tabular}{@{}cccc@{}}
\hline
\hline
                   & SVR    & GBRT   & MLP    \\
\hline
G-features               & 2.12\%          & 2.29\%          & 2.25\%          \\
A-features               & 2.43\%          & 2.65\%          & 2.42\%          \\
S-features                 & 2.63\%          & 2.58\%          & 2.55\%          \\
G-features and A-features & 2.07\%          & 2.21\%          & 2.17\%          \\
G-features and S-features  & 1.97\%          & 1.98\%          & 1.91\%          \\
A-features and S-features   & 2.39\%          & 2.31\%          & 2.31\%          \\
MSF                   & \textbf{1.89\%} & \textbf{1.82\%} & \textbf{1.78\%} \\
\hline
\hline
\end{tabular}
\end{table}
As shown in Table \ref{tab_03}, compared with models based on A-features and S-features, models based on G-features can obtain the highest forecasting accuracy. Generally speaking, the performance of models based on A-features is slightly better than those based on S-features. According to Table \ref{tab_03}, the combination of G-features and S-features could obtain a satisfactory forecasting accuracy, which is widely used in many research. While when combining the features from all three aspects, a higher forecasting accuracy can be obtained. The results indicate that introducing A-features is a great complement to G-features and S-features. 

\begin{figure*}[!t]
\centering
\includegraphics[width = 6in]{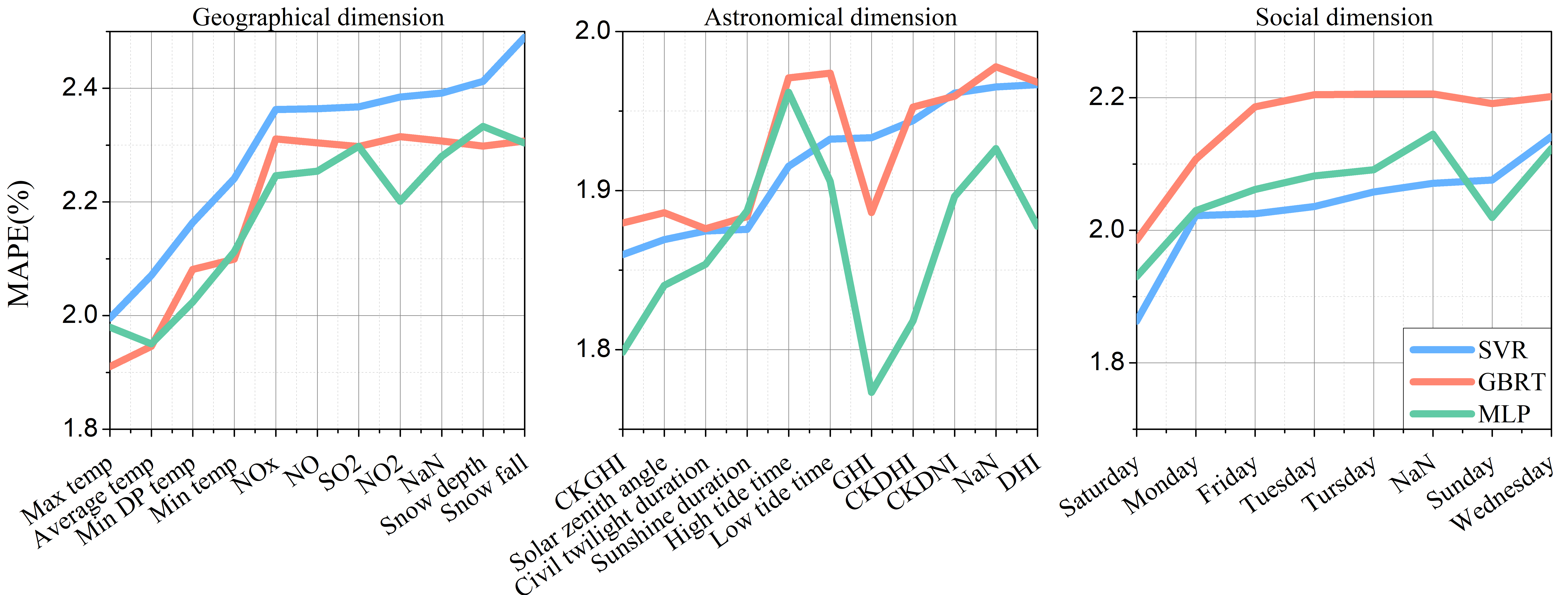}
\caption{Forecasting accuracy obtained by each feature when combined with historical load features.}
\label{fig_06}
\end{figure*}

To further analyze the importance of different features from each aspect, top 10 G-features, top 10 A-features, and top 7 S-features are used one by one with historical load features as the input and the accuracy obtained by them are shown in Fig.\ref{fig_06}
In Fig.\ref{fig_06}, NaN indicates that no corresponding feature is used. According to the results, among all G-features, temperature is the most significant feature that affects the forecasting accuracy. 
As for A-features, CKGHI, SZA, CTD, and GHI are the most important ones that improve the forecasting accuracy.
These features represent the degree of solar irradiance received by the earth and the positional relationship between the earth and the sun, which are closely related to the temperature, weather, climate, etc. Therefore, they are related to the variation of electric load.
High tide time and low tide time are related to the moon. According to the results, they have no significant effect on improving the forecasting accuracy.
For S-features, the improvement of accuracy brought by Saturday and Monday is significant. The reason behind this may be that Saturday and Monday are the first day of weekend and working days, and the power load on these two days change significantly from the previous day.

\subsection{The Correlation between Dominant Factors and Load Variation}
Factors corresponding to dominant features are dominant factors that affect the load variation. To further study the relationship between dominant factors and load variation and how they affect the forecasting result, we apply partial dependence plot (PDP) to show the correlation between dominant factors and forecasting load values.

PDP shows the response of a trained model to a single feature\cite{Kapelner2016bartMachine}. Assuming that we are studying the influence of the \emph{j}th feature, the partial dependence is defined as:
\begin{equation}
\hat f({x_j}) = \frac{1}{n}\sum\limits_{i = 1}^n {\hat f({x_j},{x_{ - j}},i)} 
\label{eq_08}
\end{equation}
where, \emph{\^{f}} represents the trained model, \emph{n} represents the number of samples in the training set, \emph{x$_{-j}$} represents all the features except for \emph{x$_{j}$}. The PDP of \emph{x$_{j}$} is defined as the average value of \emph{\^{f}} when \emph{x$_{j}$} is fixed and \emph{x$_{-j}$} varies over its marginal distribution.

\begin{figure}[!t]
\centering
\includegraphics[width = 3.3in]{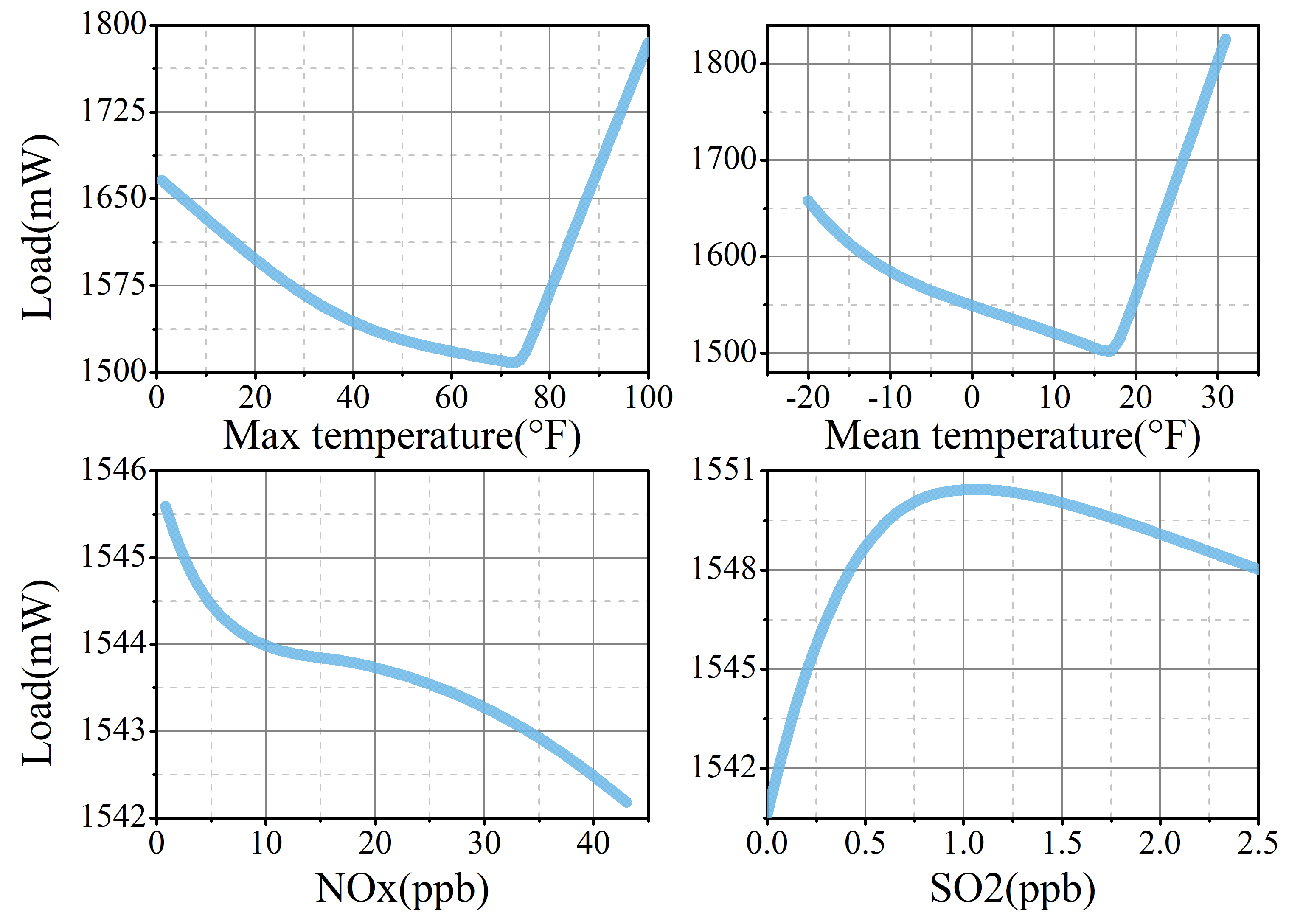}
\caption{PDP of maximum temperature, average temperature, NO$_{x}$ content, and SO$_{2}$ content.}
\label{fig_07}
\end{figure}

Fig.\ref{fig_07} shows the PDP of four G-factors, namely maximum temperature, average temperature, NO$_{x}$ content, and SO$_{2}$ content.
As shown in Fig.\ref{fig_07}, when the temperature is higher than a threshold, the load consumption is positively related to the temperature. When the temperature is lower than the threshold, the relationship between them becomes negative. This temperature threshold is called temperature balance point\cite{AHMED2018load}. The balance point is not necessarily the same in different locations. In this case, the balance point of maximum temperature is around $70^\circ$F and that of average temperature is around $15^\circ$F.
The reason behind this is that, with the increase or decrease of temperature, more air conditioners or heaters are used, as a result the load consumption increases accordingly. 
From Fig.\ref{fig_07}, we can see how the NO$_{x}$ and SO$_{2}$ content affect the forecasting result. The forecasting load value decreases with the increase of NO$_{x}$ content. When the content of SO$_{2}$ increases from 0ppb to 1ppb, the load value increases and after that, it decreases slowly.

\begin{figure}[!t]
\centering
\includegraphics[width = 3.3in]{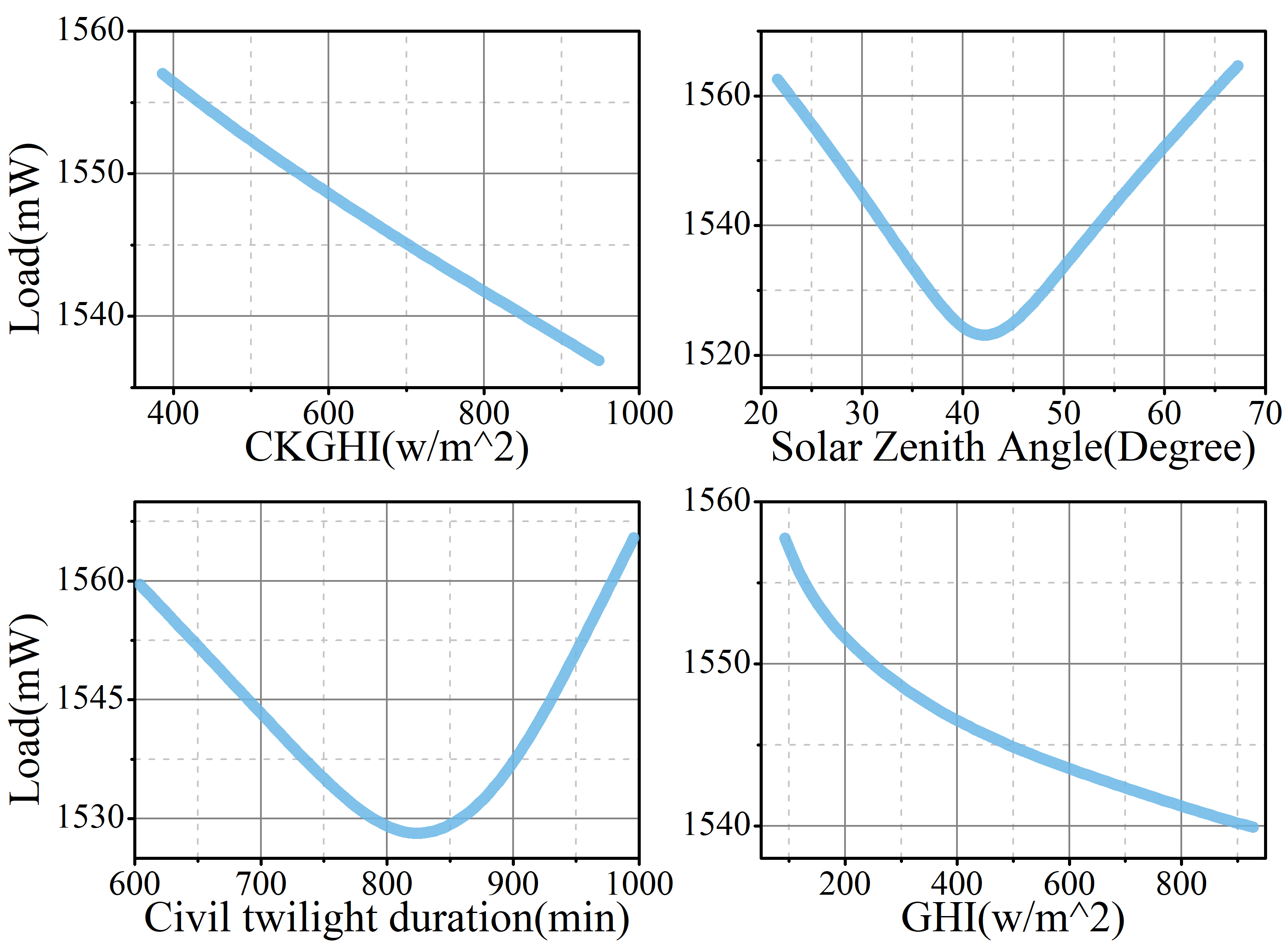}
\caption{PDP of CKGHI, SZA, CTD, and GHI.}
\label{fig_08}
\end{figure}

\begin{figure}[!t]
\centering
\includegraphics[width = 2.5in]{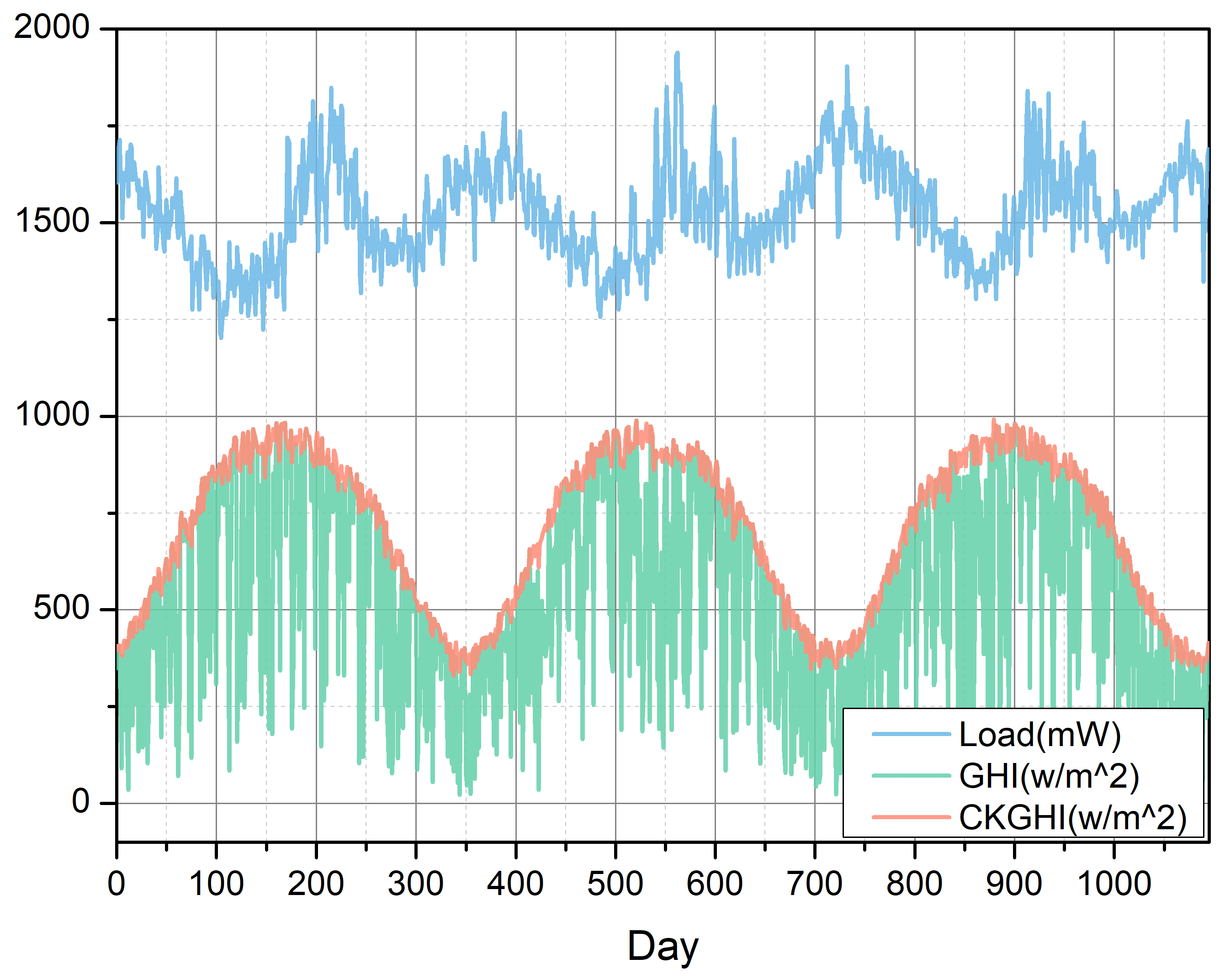}
\caption{CKGHI, GHI, and peak load of Maine from 2012 to 2014.}
\label{fig_09}
\end{figure}

\begin{figure}[!t]
\centering
\includegraphics[width = 2.5in]{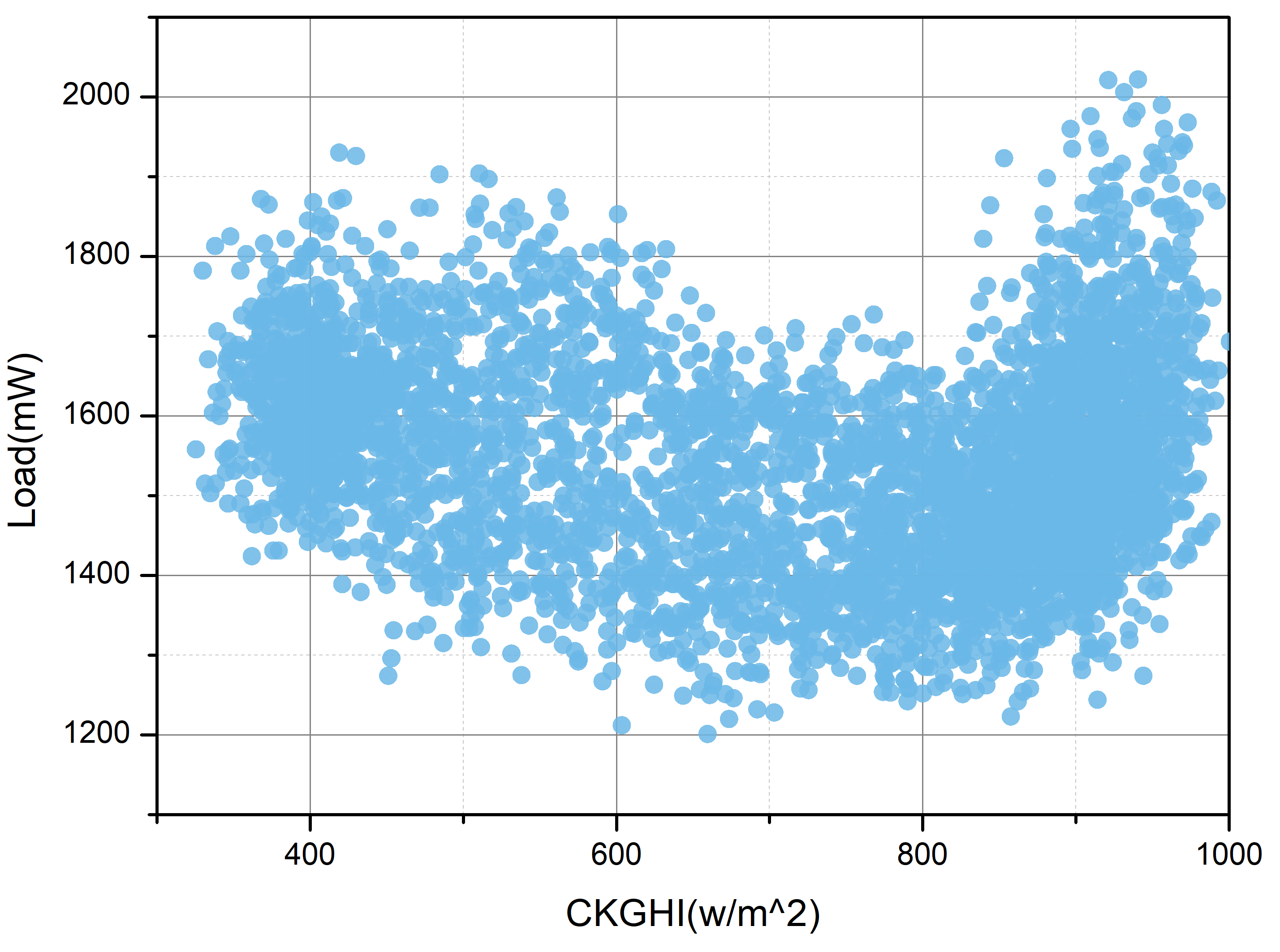}
\caption{Scatter plot of lagged CKGHI and peak load.}
\label{fig_10}
\end{figure}

Fig.\ref{fig_08} shows the PDP of four A-factors, namely CKGHI, SZA, CTD, and GHI. We can see that SZA and CTD also have a V-shape relationship between load value, showing that the balance point of SZA is around $42^\circ$ and that of CTD is around 820 minutes. The balance point of them may also vary with geographic locations. For a certain area, SZA and CTD reflect the positional relationship between the earth and the sun and they have a significant annual periodicity. Compared with using four binary variables to indicate seasons, applying SZA and CTD could better reflect the seasonal periodicity of load variation. 
CKGHI represents the total solar irradiance on a horizontal surface under clear sky condition. GHI represents the actual solar irradiance on a horizontal surface. As shown in Fig.\ref{fig_08}, CKGHI and GHI are negatively related to load value. 
Fig.\ref{fig_09} shows the CKGHI, GHI, and daily peak load of Maine from 2012 to 2014. They show an obvious annual periodicity, whereas the period of CKGHI and GHI are twice as long as that of the peak load. It can be observed that two peaks of daily load in a single year have a certain correspondence with the peak and valley of CKGHI and there is a phase difference between them. The peak load lags behind the CKGHI by about 50 days. 
Fig.\ref{fig_10} shows an obvious V-shape relationship between the peak load and 50-days-lagged CKGHI. According to geographical knowledge, due to the large specific heat capacity of the ocean, the accumulation of heat received by the earth lags behind solar irradiance. Usually, the lag of temperature behind solar radiation in the USA is around 26 to 60 days\cite{Prescott1951The}. The accumulation of heat results in affecting the temperature, ocean current, weather, etc., and further affecting the load variation. Therefore, the load variation lags behind the change of solar irradiance.

\subsection{Summary of Dominant Features}
Based on the above analysis, among all three aspects, G-features have the most important impact on improving the forecasting accuracy and temperature is the dominant feature. There is a V-shape relationship between temperature and load. In Maine state, the balance point of maximum temperature and average temperature are around $70^\circ$F and $15^\circ$F, respectively.

The influence of A-features is higher than that of S-features. Features related to the sun play an important role in improving the forecasting accuracy. Among them, the relationship between SZA, CTD, and load is not linear, showing a V-shape pattern and the balance point of them are around $42^\circ$ and 820 minutes. CKGHI and GHI represent the solar radiation received by the earth, the period of them is twice as long as that of load variation. There is a V-shape relationship between load and lagged CKGHI.

Among the top 7 S-features, the importance of Saturday and Monday is more significant. The reason behind may be that Saturday and Monday are the first days of weekend and working days, and human activities on these days change significantly and thus affecting the load variation pattern.

\section{Conclusion}
Based on MSF, this paper studies the dominant factors that affect the load variation and proposes a short-term load forecasting method. The proposed method collects up to 80 features from astronomical, geographical, and social aspects to construct MSF. The features selected based on MSF provide the forecasting model with diversity and large-scale data support and finally improve the forecasting accuracy. This research has revealed that G-features have the most significant impact on improving the forecasting accuracy, in which temperature is the dominant feature that improves forecasting accuracy. The influence of A-features is more significant than that of S-features and features related to the sun have a more obvious effect on improving the accuracy, which is obviously ignored in previous research. Among all S-features, Saturday and Monday are the most important ones for load forecasting. Among all the dominant factors, temperature, SZA, and CTD have a V-shape relationship with the load. There is a V-shape relationship between lagged CKGHI and load, and a negative linear correlation between GHI and load.

Case studies are carried out based on the real-world datasets, including the daily peak load of Maine and Texas, and the half-hourly load of NSW. Since these datasets are from different regions, where the climate, residents’ living habits, weather, geographical environment are totally different from each other. The experiment results based on these datasets show that the proposed MSF is dataset-independent.
Moreover, case studies show that the forecasting performance of different learning algorithms would be improved with the application of MSF, indicating that the proposed MSF is model-independent.
The research conducted in this paper demonstrates the wide scope of applications and strong generalization ability of MSF.

With the development of the smart grid and the applications of the Internet of Things in power systems, more and more data from different aspects can be obtained in the future and the complex factor-load nexus will be further studied.

\ifCLASSOPTIONcaptionsoff
  \newpage
\fi



\bibliographystyle{IEEEtran}
\bibliography{Manuscipt}

\begin{thebibliography}{10}
\providecommand{\url}[1]{#1}
\csname url@samestyle\endcsname
\providecommand{\newblock}{\relax}
\providecommand{\bibinfo}[2]{#2}
\providecommand{\BIBentrySTDinterwordspacing}{\spaceskip=0pt\relax}
\providecommand{\BIBentryALTinterwordstretchfactor}{4}
\providecommand{\BIBentryALTinterwordspacing}{\spaceskip=\fontdimen2\font plus
\BIBentryALTinterwordstretchfactor\fontdimen3\font minus
  \fontdimen4\font\relax}
\providecommand{\BIBforeignlanguage}[2]{{%
\expandafter\ifx\csname l@#1\endcsname\relax
\typeout{** WARNING: IEEEtran.bst: No hyphenation pattern has been}%
\typeout{** loaded for the language `#1'. Using the pattern for}%
\typeout{** the default language instead.}%
\else
\language=\csname l@#1\endcsname
\fi
#2}}
\providecommand{\BIBdecl}{\relax}
\BIBdecl

\bibitem{Hverstad2015short}
B.~A. {Høverstad}, A.~{Tidemann}, H.~{Langseth}, and P.~{Öztürk},
  ``Short-term load forecasting with seasonal decomposition using evolution for
  parameter tuning,'' \emph{IEEE Transactions on Smart Grid}, vol.~6, no.~4,
  pp. 1904--1913, 2015.

\bibitem{xie2016relative}
J.~Xie, Y.~Chen, T.~Hong, and T.~D. Laing, ``Relative humidity for load
  forecasting models,'' \emph{IEEE Transactions on Smart Grid}, vol.~9, no.~1,
  pp. 191--198, 2016.

\bibitem{wi2011holiday}
Y.-M. Wi, S.-K. Joo, and K.-B. Song, ``Holiday load forecasting using fuzzy
  polynomial regression with weather feature selection and adjustment,''
  \emph{IEEE Transactions on Power Systems}, vol.~27, no.~2, pp. 596--603,
  2011.

\bibitem{wang2016electric}
P.~Wang, B.~Liu, and T.~Hong, ``Electric load forecasting with recency effect:
  A big data approach,'' \emph{International Journal of Forecasting}, vol.~32,
  no.~3, pp. 585--597, 2016.

\bibitem{zeng2017peak}
P.~Zeng and M.~Jin, ``Peak load forecasting based on multi-source data and
  day-to-day topological network,'' \emph{IET Generation, Transmission \&
  Distribution}, vol.~12, no.~6, pp. 1374--1381, 2017.

\bibitem{son2017short}
H.~Son and C.~Kim, ``Short-term forecasting of electricity demand for the
  residential sector using weather and social variables,'' \emph{Resources,
  conservation and recycling}, vol. 123, pp. 200--207, 2017.

\bibitem{pan2019learning}
Z.~Pan, C.~Sheng, and J.~Min, ``A learning framework based on weighted
  knowledge transfer for holiday load forecasting,'' \emph{Journal of Modern
  Power Systems and Clean Energy}, vol.~7, no.~2, pp. 329--339, 2019.

\bibitem{guo2018deep}
Z.~Guo, K.~Zhou, X.~Zhang, and S.~Yang, ``A deep learning model for short-term
  power load and probability density forecasting,'' \emph{Energy}, vol. 160,
  pp. 1186--1200, 2018.

\bibitem{moral2017integrating}
J.~Moral-Carcedo and J.~P{\'e}rez-Garc{\'\i}a, ``Integrating long-term economic
  scenarios into peak load forecasting: An application to spain,''
  \emph{Energy}, vol. 140, pp. 682--695, 2017.

\bibitem{Jiang2018Ashort}
H.~{Jiang}, Y.~{Zhang}, E.~{Muljadi}, J.~J. {Zhang}, and D.~W. {Gao}, ``A
  short-term and high-resolution distribution system load forecasting approach
  using support vector regression with hybrid parameters optimization,''
  \emph{IEEE Transactions on Smart Grid}, vol.~9, no.~4, pp. 3341--3350, 2018.

\bibitem{Wang2011ELITE}
B.~Wang and H.-D. Chiang, ``Elite: Ensemble of optimal input-pruned neural
  networks using trust-tech,'' \emph{IEEE Transactions on Neural Networks},
  vol.~22, no.~1, pp. p.96--109, 2011.

\bibitem{Wang2018an}
Y.~{Wang}, Q.~{Chen}, M.~{Sun}, C.~{Kang}, and Q.~{Xia}, ``An ensemble
  forecasting method for the aggregated load with subprofiles,'' \emph{IEEE
  Transactions on Smart Grid}, vol.~9, no.~4, pp. 3906--3908, 2018.

\bibitem{nowotarski2016improving}
J.~Nowotarski, B.~Liu, R.~Weron, and T.~Hong, ``Improving short term load
  forecast accuracy via combining sister forecasts,'' \emph{Energy}, vol.~98,
  pp. 40--49, 2016.

\bibitem{zhou2017holographic}
M.~Zhou and M.~Jin, ``Holographic ensemble forecasting method for short-term
  power load,'' \emph{IEEE Transactions on Smart Grid}, vol.~10, no.~1, pp.
  425--434, 2017.

\bibitem{jingrui2018load}
X.~Jingrui and H.~Tao, ``Load forecasting using 24 solar terms,'' \emph{Journal
  of Modern Power Systems and Clean Energy}, vol.~6, no.~2, pp. 208--214, 2018.

\bibitem{alipour2019assessing}
P.~Alipour, S.~Mukherjee, and R.~Nateghi, ``Assessing climate sensitivity of
  peak electricity load for resilient power systems planning and operation: A
  study applied to the texas region,'' \emph{Energy}, vol. 185, pp. 1143--1153,
  2019.

\bibitem{Wang2019Bi}
S.~Wang, X.~Wang, S.~Wang, and D.~Wang, ``Bi-directional long short-term memory
  method based on attention mechanism and rolling update for short-term load
  forecasting,'' \emph{International journal of electrical power and energy
  systems}, vol. 109, no. JUL., pp. 470--479, 2019.

\bibitem{chipman2010bart}
H.~A. Chipman, E.~I. George, R.~E. McCulloch \emph{et~al.}, ``Bart: Bayesian
  additive regression trees,'' \emph{The Annals of Applied Statistics}, vol.~4,
  no.~1, pp. 266--298, 2010.

\bibitem{Kapelner2016bartMachine}
A.~Kapelner and J.~Bleich, ``bartmachine: Machine learning with bayesian
  additive regression trees,'' \emph{Journal of Statistical Software}, vol.
  070, no.~4, 2016.

\bibitem{AHMED2018load}
T.~Ahmed, D.~Vu, K.~Muttaqi, and A.~Agalgaonkar, ``Load forecasting under
  changing climatic conditions for the city of sydney, australia,''
  \emph{Energy}, vol. 142, pp. 911 -- 919, 2018.

\bibitem{Prescott1951The}
J.~A. Prescott and J.~A. Collins, ``The lag of temperature behind solar
  radiation,'' \emph{Quarterly Journal of the Royal Meteorological Society},
  vol.~77, no. 331, pp. 121--126, 1951.

\end{thebibliography}
%



%





\end{document}